\documentclass[letterpaper, 10 pt, conference]{RO-MAN26/ieeeconf}  
\IEEEoverridecommandlockouts
\usepackage{cite}
\usepackage{amsmath,amssymb,amsfonts}
\usepackage{algorithmic}
\usepackage{graphicx}
\usepackage[hyphens,spaces,obeyspaces]{url}
\usepackage{hyperref}
\usepackage{todonotes}
\usepackage{textcomp}
\usepackage{xcolor}
\usepackage{bm}
\usepackage{threeparttable}
\usepackage{siunitx}
\usepackage{multirow}
\usepackage{subfig}
\usepackage{comment}
\usepackage{tikz}
\usepackage{url}

\usepackage{caption}
\usepackage{hyperref}
\usepackage{lipsum}
\usepackage{gensymb}
\usetikzlibrary{shapes}

\usepackage[skip=8pt,font=small]{caption}

\def\BibTeX{{\rm B\kern-.05em{\sc i\kern-.025em b}\kern-.08em
    T\kern-.1667em\lower.7ex\hbox{E}\kern-.125emX}}

\hypersetup{
	colorlinks=true,
	linkcolor=black,
	filecolor=magenta,      
	urlcolor=black,
}

\begin{document}

\title{\LARGE \bf Legible and Intuitive Multi-modal Robot State and Intent Communication Validated in Online and Real-world Studies
}

\author{Tim~Schreiter$^{1,3}$,
       Jens V. Rüppel$^1$,
       Andrey Rudenko$^1$, 
       Martin Magnusson$^2$,
       and~Achim J.~Lilienthal$^{1,2}$
\thanks{$^{1}$Chair of Perception for Intelligent Systems, Munich Institute of Robotics and Machine Intelligence (MIRMI), Technical University of Munich (TUM), Germany 
{\tt\small \{tim.schreiter, jens.v.rueppel, andrey.rudenko, achim.j.lilienthal\}@tum.de}}%
\thanks{$^{2}$Centre for Applied Autonomous Sensor Systems (AASS),
	\"Orebro University, Sweden {\tt\small martin.magnusson@oru.se}}
    \thanks{$^{3}$Robotics Institute Germany (RIG)}
\thanks{This work was supported by the European Union’s Horizon 2020 research and innovation program under grant agreement No. 101017274 (DARKO)}}

\maketitle

\begin{abstract}
Effective robot-to-human communication can increase transparency and trust, reduce uncertainty, and contribute to safer collaboration in shared workspaces. Designing and validating an effective robot communication strategy is challenging due to the varying and often limited communication modalities across robots, differences in how diverse recipients interpret messages, and the underexplored virtual-to-real gap in studies of communication legibility. We present a systematic, large-scale comparative validation of existing communication strategies for a mobile non-humanoid robot across message types and settings (online and in-person). Based on the prescribed message types in the existing standards for industrial robots, we realize and compare a low-expressive, unimodal LED-based strategy with a highly expressive, multimodal one that leverages robotic gaze, gestures, and voice. For each strategy, we analyze the communication of a turning intention, an attention request, error status, whether the robot is stuck, and whether it is functioning normally. We evaluate these strategies in replicated online and in-person experiments. We find strong evidence that highly expressive multimodal communication is perceived as more legible and intuitive than unimodal LED-based communication. Comparing the online and real-world study findings, we observe a notable decrease in overall legibility, particularly for signaling with LEDs. Similarly, confidence in message interpretation decreases during the real-world evaluation.
\end{abstract}

\section{Introduction}

Although robot-to-human communication is important for ensuring safe and intuitive interaction in shared spaces, current industrial standards (e.g., ISO 10218 and ANSI/ITSDF B56.5-2019) offer little guidance on how to convert essential information into legible signals \cite{schott2023literature}. Prior work has explored a wide range of robots and external Human-Machine Interfaces (eHMI), including displays, through various online and in-person experiments that reflect different levels of expressive bandwidth \cite{brill2023external, rouchitsas2019external}. On one side of the expressive bandwidth spectrum, there are simple visual cues, such as lights or projections, which are robust but abstract \cite{baraka2016enhancing}. On the other side, there are complex multimodal signals that combine speech and expressive nonverbal cues, perceived as more natural and capable of conveying more complex messages for improved coordination, but which also increase system complexity \cite{schreiter2023advantages, akalin2022you}. This opens up a vast design space for developing new robot communication strategies. However, due to the large design space, there remains a lack of comparative research assessing the effectiveness of different levels of expressive bandwidth, especially in realistic, ``in the wild'' environments \cite{tanjim2025help, esterwood2025virtually}.

We address this gap by systematically comparing two existing communication strategies ($\mathcal{S}$) at opposite ends of the expressive bandwidth spectrum. One strategy is a unimodal communication strategy with low expressive bandwidth, relying on light-based signaling with LEDs. The other one is a multimodal communication strategy with higher expressive bandwidth, leveraging a combination of nonverbal and verbal robotic cues. We implement the strategies on a mobile, non-humanoid service robot and conduct an online survey and an in-person experiment at a science museum that replicates it. In these two contexts, using both $\mathcal{S}$, we modulate five different messages and study perceived legibility and intuitiveness of robot communication in shared workspaces.

\begin{figure}[!t]
    \centering
    \includegraphics[trim={0cm 5cm 0cm 2.5cm}, clip, width=\columnwidth]{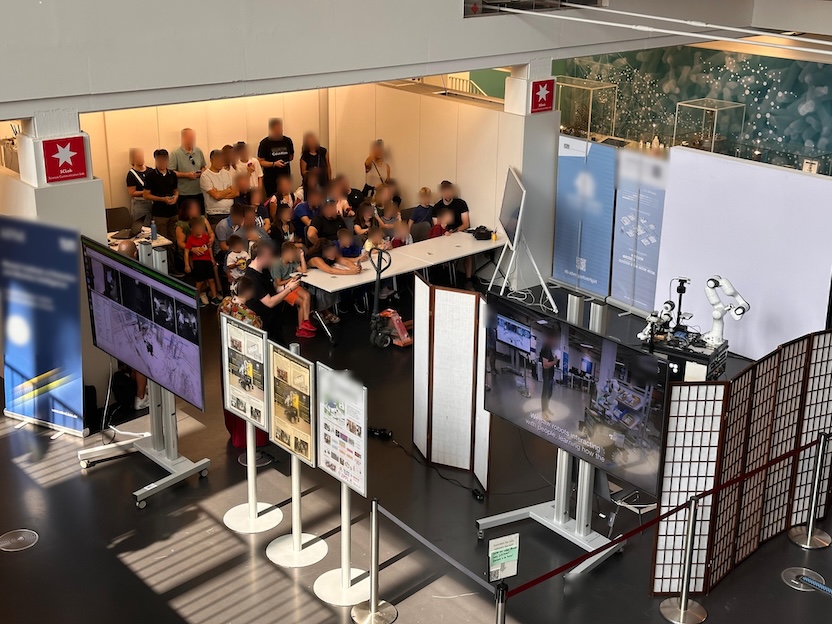}
    \caption{Visitors of a museum watching a robot communicating different messages during our real-world experiment.}
    \label{fig:frontal}
\end{figure}

In this paper, we study two research questions: \textbf{RQ1} How does the physical setting (online video viewing vs. real-world interaction) affect the perceived effectiveness of robot communication across message types? \textbf{RQ2} How do perceived legibility and intuitiveness differ between communication strategies with low or high expressive bandwidth?

We use a mixed-methods approach comprising an online survey (N=100) using video-based scenarios and a follow-up in-person validation study (N=139) at a public science museum. Following the formalisms of Dragan et al. \cite{dragan2013legibility} and Lichtenthaler et al. \cite{lichtenthaler2016legibility}, we evaluate these interactions through the lenses of legibility (the ability to infer intent) and intuitiveness (the ease of interpretation). We hypothesize that, while real-world environmental factors will attenuate communication effectiveness compared to online settings (H1), strategies with higher expressive bandwidth will significantly outperform those with lower bandwidths for complex, action-oriented messages such as attention requests and navigation turns (H2). We support our analysis with sub-hypotheses outlined in H1.1 to quantify the legibility drop and H2.1-H2.3 to analyze the effect on individual messages.

Our study provides a systematic evaluation of service robot communication by bridging the gap between experiencing the robot on a screen and in-person. By comparing a unimodal communication strategy with low expressive bandwidth ($\mathcal{S}_{low}$) with a multimodal strategy with high expressive bandwidth ($\mathcal{S}_{high}$), we demonstrate that low-bandwidth cues are effective for communicating simple, safety-critical internal states such as system errors. Conversely, due to increased clarity or non-ambiguity, $\mathcal{S}_{high}$ significantly increases perceived legibility and intuitiveness for complex, spatially anchored messages such as turn announcements, where explicit multimodal cues help resolve inherent ambiguities. Our findings clearly show a critical virtual-to-real gap, highlighting the need to validate novel HRI concepts not only via online video viewing but also with a physically present robot. Our findings thus suggest a hybrid design paradigm to facilitate coordinated human-robot interaction in shared spaces. We recommend using robust, unimodal cues with low expressive bandwidth for system status and combining multiple modalities to provide high expressive bandwidth for intent signaling.

\section{Related Work}


A challenge for any autonomous agent that operates in spaces shared with humans is to achieve mutual predictability through transparent communication of states and intents \cite{schott2023literature}. Explicit communication by the agent through cues such as motion, lighting, or other visual signals can provide the necessary transparency and interpretability for interactions \cite{may2015show, fernandez2018passive, lemasurier2021methods}. However, the real challenge is to achieve legibility (correct interpretation) and intuitiveness (ease of understanding). We adopt operational definitions for these key terms used throughout this paper, e.g., the American Psychological Association's definition for \textit{intent\footnote{\url{https://dictionary.apa.org/intention}}} as a prior decision to perform a goal-directed action. In our experiment, the intent of the robot is communicated through explicit messages preceding an upcoming action. In line with prior studies on legibility in human–robot interaction \cite{dragan2013legibility, lichtenthaler2016legibility}, we operationalize \textit{legibility} as the extent to which an observer can correctly interpret the meaning of a robot’s communicated message and \textit{intuitiveness} as the perceived ease with which a communicated message can be understood without prior instruction \cite{van2025seeing}. These terms guide our understanding of how the communication of the robot should be designed.

\subsection{Communication Strategies for Non-Humanoid Robots}
Following a framework of intentional design \cite{kunold2022framework}, robot communication strategies are purposefully modulated outputs that span the Expressive Bandwidth (EB) spectrum. Even basic robotic platforms offer multiple modalities, creating an expansive design space \cite{kunold2022framework}, where channel selection significantly impacts transparency of the interaction \cite{schott2023literature}. At the lower end, unimodal light-based strategies are valued for robustness, cost-effectiveness, and visibility in safety-critical industrial environments \cite{fernandez2018passive, lemasurier2021methods, thiem2022verfahren}. While effective for communicating basic status and motion intent \cite{baraka2018mobile, thiem2022verfahren, baraka2016enhancing, bacula2023integrating}, their abstract and often arbitrarily coded nature can lead to ambiguity, requiring contextual grounding and symbolic interpretation \cite{fernandez2018passive, lemasurier2021methods}. On the other hand, high-EB strategies combine multiple modalities, such as verbal and non-verbal cues, to achieve greater semantic specificity (clarity and lack of ambiguity in the associated meaning) \cite{prati2013semantic}. By leveraging iconic or intrinsically coded cues that are immediately comprehensible in a redundant mix of modalities, high-EB strategies can reduce human uncertainty and improve coordination during complex navigation or recovery tasks \cite{akalin2025explain, halilovic2024exploring, schreiter2023advantages}. The characteristics of these EB extremes illustrate the need for comparative studies to better weigh their functional trade-offs when designing novel, ideally intuitive communication strategies \cite{schott2023literature, esterwood2025virtually}.

The success of a communication strategy can be viewed as a function of the semantic specificity required by an individual message \cite{prati2013semantic}. How effectively a receiver decodes the message of a robot is tied to its morphological affordances and the semantic complexity available for encoding \cite{tanjim2025help, schott2023literature}. For messages with low semantic complexity, such as binary status updates or battery-level indicators, unimodal low-bandwidth channels, such as LED arrays, are effective and minimize cognitive load \cite{baraka2016enhancing, akalin2025explain}. However, as message content reaches a certain complexity threshold, like conveying a precise navigational goal, these abstract messages become prone to misinterpretation due to a lack of semantic specificity \cite{tanjim2025help, bhattathiri2024unlocking}. In such scenarios, high-bandwidth multimodal strategies can combine speech and deictic gestures to resolve spatial ambiguity and provide explicit cues necessary for coordination \cite{akalin2025explain, brill2023external, halilovic2024exploring}.  A robot’s physical appearance implies expectations for a communication partner. Social features like a head or eyes automatically trigger assumptions about social intelligence and communication competence \cite{breazeal2016social, roesler2023embodiment}. Kunold and Onnasch \cite{kunold2022framework} argue that because humans assign constant meaning to all observed behaviors, a robot ``can not not communicate \cite{watzlawick2011pragmatics}'', making it critical for developers to be intentional about design choices. If carefully selected, each message achieves specific effects on the receiver, enabling researchers, in turn, to improve task coordination and elicit the appropriate human response, ensuring that even unintentional cues do not lead to communicative failures \cite{akalin2025explain}.

\subsection{Online versus In-Person Evaluation in HRI}
Online and virtual evaluations offer high scalability to gain early design insights, providing insights into the controlled laboratory where the design takes place \cite{esterwood2025virtually}, without the logistical burden of bringing in participants. However, findings might not be transferable to real-world interactions, where a physically embodied robot enables spatial and temporal proximity \cite{breazeal2016social, rae2013body} or an improvement in objective metrics, such as reaction times and joint attention \cite{schreiter2023advantages, roesler2023embodiment}. While the ``Embodiment Hypothesis’’ suggests that physical co-location also enhances trust and coordination \cite{AdmoniReview, breazeal2016social}, recent meta-analyses indicate that these differences can be statistically subtle and context dependent \cite{esterwood2025virtually}. A major challenge remains the ``virtual-to-real gap'' \cite{esterwood2025virtually}. Video-based methods often strip away real-world stimuli such as motor noise or the ``psychological weight’’ of embodied presence, which are essential to a user's impression of a robot \cite{fiorini2024role}. Real-world environments also introduce both sensory and cognitive ``noise,'' such as sounds or divided attention (distractions),  degrading message interpretability and potentially leading to communication failures \cite{bhattathiri2024unlocking, tanjim2025help, mahadevan2018communicating}. Consequently, ``in-the-wild’’ validation in public spaces is required to establish if a robot can communicate legibly and intuitively in the noisy reality of spaces shared with humans \cite{AdmoniReview, breazeal2016social}.

\section{Methods}\label{sec:methods}

We examine the perceived legibility and intuitiveness of robot intent communication through a two-stage evaluation process. First, we conduct an online survey (see Section \ref{subsec:OnlineSurvey}) followed by a close replication of users' experiences in a real-world setting (see Section \ref{subsec:Museum}). Using a Robotnik RB-Kairos base equipped with a built-in LED stripe and an additional interface character mounted on the base (ARMoD \cite{schreiter2023advantages}), we implement two communication strategies $\mathcal{S}$ with different expressive bandwidths. The first one, $\mathcal{S}_{low}$, is a unimodal communication strategy with low expressive bandwidth that leverages the LED stripes of the mobile base for communication. The second one, $\mathcal{S}_{high}$, is a high-expressive multimodal strategy that combines the ARMoD's (NAO) multiple communication channels, such as robotic gestures, head orientation, and speech. For each $\mathcal{S}$, we design five messages (see Section \ref{subsec:msg}). We study their impact on participants' perceived legibility and intuitiveness in both of our contexts, using a set of hypotheses. \textbf{Hypothesis 1:} Communication effectiveness differs between online and real-world settings, specifically, (\textbf{H1.1:}) Perceived legibility and intuitiveness are reduced in real-world settings due to environmental and attentional factors. \textbf{Hypothesis 2:} Communication strategies with higher expressive bandwidth are perceived as more legible and more intuitive than low-expressivity visual (LED-based) signaling for selected message types, specifically \textbf{H2.1:} for attention
requests, \textbf{H2.2:} for error state communication, and \textbf{H2.3:} for turn announcements. Finally, we conduct an exploratory analysis of speech as a standalone semantic cue in $\mathcal{S}_{high}$, a first step in isolated communication factor analysis, planned in future work.

\subsection{Communicated Messages}\label{subsec:msg}

\begin{figure}[!t]
    \vspace{0.2cm}
    \centering
    
    \subfloat[\textbf{Idle} \textit{Left:} NAO sitting in power saving (idle) mode \textit{Right:} LEDs flashing in a steady green]{
        \label{fig:row_idle}
        \includegraphics[keepaspectratio, trim={0cm 4.9cm 0cm 0cm}, clip, width=0.45\linewidth]{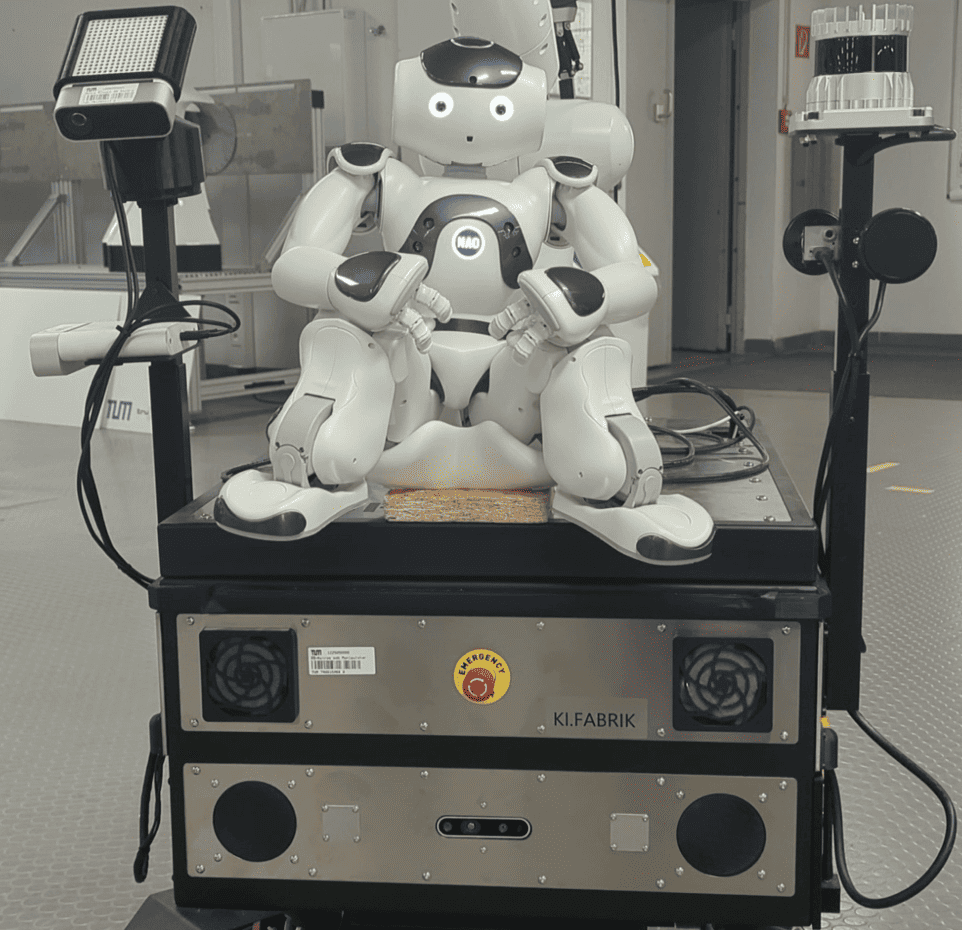}%
        \hspace{0.2cm}
        \includegraphics[height=3.8cm, trim={0cm 3cm 0cm 3cm}, clip, keepaspectratio, width=0.45\linewidth]{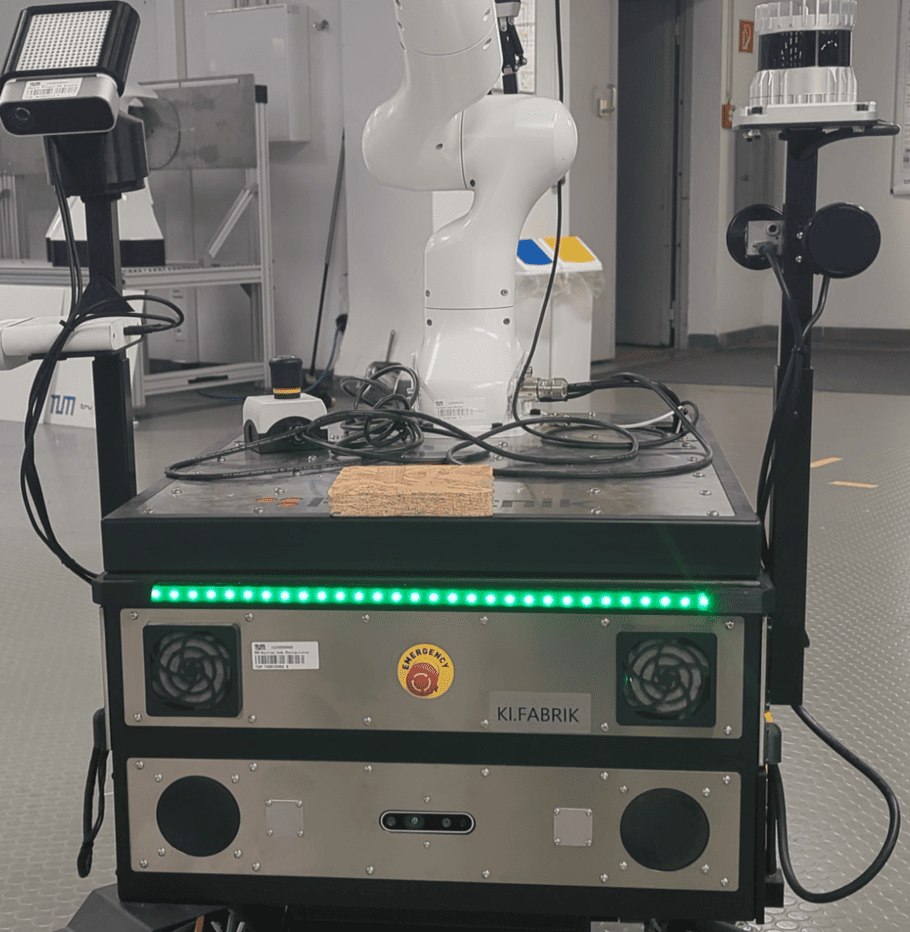}
    }  \vspace{-0.1cm}

    \subfloat[\textbf{Attention}  \textit{Left:} NAO waving arm in combination with verbal utterance \textit{Right:} LEDs deploying a white pulsating light (2 Hz)]{
        \label{fig:row_attention}
        \includegraphics[height=1.8cm,width=0.45\linewidth]{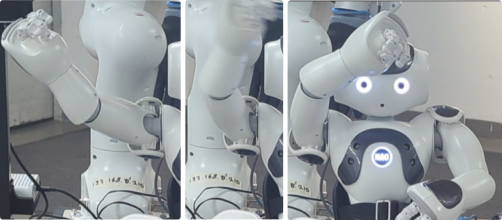}%
        \hspace{0.2cm}
        \includegraphics[height=1.8cm, width=0.45\linewidth]{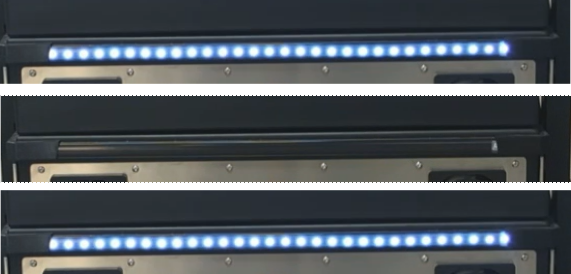}
    }
    \vspace{-0.1cm}
    
    \subfloat[\textbf{Turn} \textit{Left:} NAO cueing the turning direction using gaze, speech, and pointing. the \textit{Right:} LEDs deploying yellow walking light (1.43 Hz).]{
        \label{fig:row_turn}
        \includegraphics[height=2.4cm, width=0.45\linewidth]{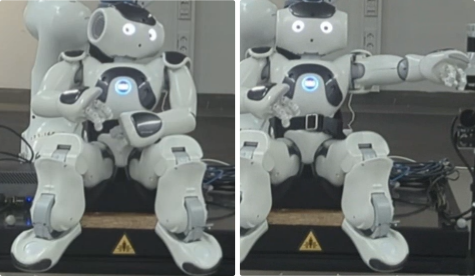}%
        \hspace{0.2cm}
        \includegraphics[height=2.4cm, width=0.45\linewidth]{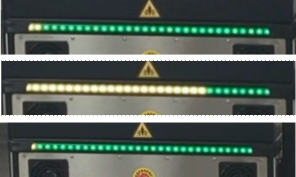}
    }
    
    \subfloat[\textbf{Error} \textit{Left:} NAO shakes head, uses a verbal utterance and a predifned gesture \textit{Right:} LEDs deploy a red pulsating light (1 Hz).]{
        \label{fig:row_error}
        \includegraphics[height=2.8 cm, width=0.45\linewidth]{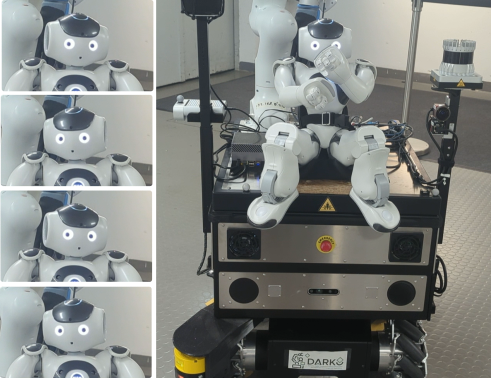}%
        \hspace{0.2cm}
        \includegraphics[height=2.8 cm, width=0.45\linewidth]{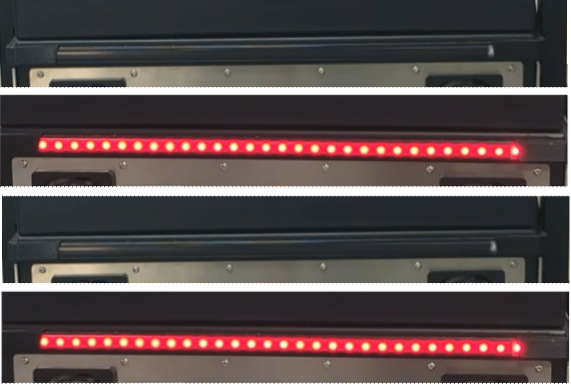}
    }
    \vspace{-0.1cm} 

    \subfloat[\textbf{Obstacle} \textit{Left:} NAO using a deictic gesture and verbal utterance \textit{Right:} LEDs flash a constant red light at the obstacle.]{
        \centering
        \label{fig:row_obst}
        \includegraphics[keepaspectratio, width=0.45\linewidth]{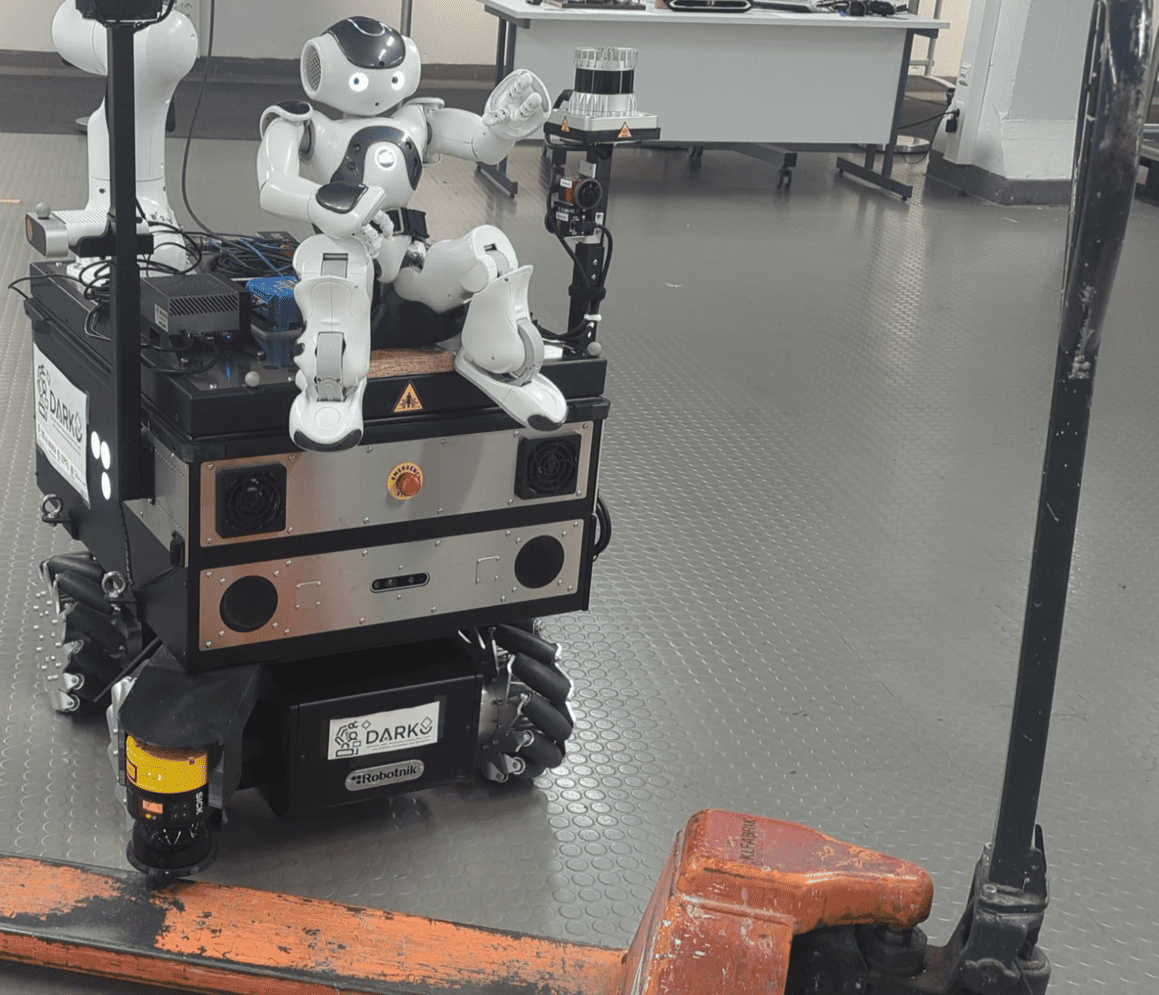}%
        \hspace{0.6cm}
        \includegraphics[height=3.3cm, keepaspectratio, width=0.45\linewidth]{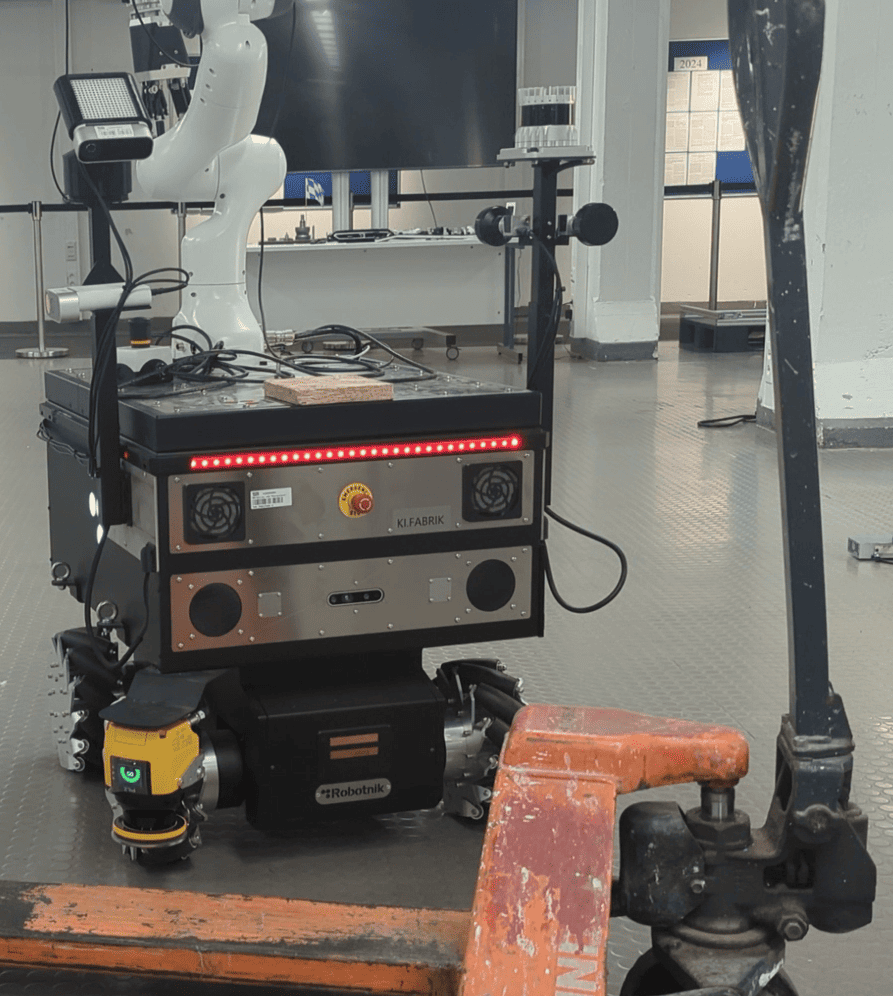}
        \hspace{0.2cm}
    }
    
    \caption{Messages are encoded in one of two communication strategies $\mathcal{S}$, varying in expressive bandwidth (High or Low). \textit{Left} $\mathcal{S}_{high}$ using a NAO \textit{Right}  $\mathcal{S}_{low}$ using an LED stripe}
    \label{fig:NAOvsDARKO_1}
\end{figure}

We define five core messages to communicate status and intentions for a mobile service robot in shared public spaces. Each message is realized through one of two distinct strategies $\mathcal{S}_{low}$ or $\mathcal{S}_{high}$) All messages are illustrated in Figure ~\ref{fig:NAOvsDARKO_1}. During experimental trials, robot communication was supervised via a remote controller, allowing an experimenter to trigger specific cues at predefined intervals. We provide a supplemental video playlist for all messages\footnote{\url{https://tinyurl.com/roman2026schrtim}}, which are the videos used in our online survey. 

\subsubsection{Idle}

This message is communicated when the robot is either between active missions or planning its next objective. For this message, we use the default behaviors inherent to each platform's morphology, as defined by its manufacturer. \textbf{$\mathcal{S}_{low}$:} The robot base displays a steady green LED pulse, a common industrial convention for ``normal operation'' or ``ready'' status \cite{baraka2016enhancing}. \textbf{$\mathcal{S}_{high}$:} The NAO adopts a predefined sitting posture to signal inactivity. While the eyes remain illuminated, they are not assigned a dynamic communicative role in this condition to isolate the effect of the static posture.

\subsubsection{Attention}

For the attention request message, we contrast a high-salience visual signal with an explicit social greeting. \textbf{$\mathcal{S}_{low}$:} We use white pulsating light for the LEDs at 2~Hz, due to demonstrated high salience and effectiveness in capturing peripheral attention \cite{baraka2016enhancing, torta2015evaluation}. \textbf{$\mathcal{S}_{high}$:} For this message, we leverage redundant and explicit cues that clearly identify the robot as the source of communication  \cite{breazeal2016social, akalin2025explain}. The NAO utilizes a multimodal approach, combining a physical waving gesture with a subsequent verbal utterance (``Excuse me. I need assistance!'').

\subsubsection{Turn} We design this message to communicate a navigation goal with high semantic complexity, requiring the human receiver to infer a specific future trajectory to avoid a potential collision. The robot drives forward, pauses, and then expresses its intention to turn left. 
We also evaluate this message in a crossing scenario in which a robot and a human approach the same point. The robot indicates a left turn just before their paths cross. \textbf{$\mathcal{S}_{low}$:} The RB-Kairos base uses a yellow light sweeping (inspired by automotive eHMI conventions for directional signaling \cite{baraka2016enhancing, bhattathiri2024unlocking}) across the LED strip from left to right (1.5 Hz) against the standard green background. \textbf{$\mathcal{S}_{high}$:} The NAO robot turns its head left, points in that direction, and announces ``Turning left.'', explicitly resolving spatial ambiguity \cite{akalin2025explain, roesler2023embodiment}. 

\subsubsection{Error}


This message communicates an internal system failure that requires immediate human awareness. \textbf{$\mathcal{S}_{low}$:} Deploys red pulsating LEDs at 1 Hz, utilizing the universal color convention for ``warning'' or ``failure'' \cite{baraka2016enhancing}. \textbf{$\mathcal{S}_{high}$:} The NAO deploys a predefined sequence of shaking its head, crossing its arms in front of its chest, and stating, ``Error, something went wrong'', to convey the robot's internal state.

\subsubsection{Obstacle}


This message represents a request for help when the robot’s path is physically obstructed, requiring human intervention to continue operating. The robot stops in front of the obstruction and rotates its base by 45$^\circ$ to improve signal visibility for the human receiver. \textbf{$\mathcal{S}_{low}$:} The robot base pulsates a red light with the LEDs at 1~Hz, similar to the Error message. \textbf{$\mathcal{S}_{high}$:} NAO looks and points directly toward the obstacle while stating: ``The way is blocked. Please clear the obstacle in front of me. I cannot move until the way is clear''. The cue is designed to identify the source of the error and list the required cooperative action, which helps eliminate ambiguities \cite{akalin2025explain, roesler2023embodiment}.

\subsection{Online Survey}\label{subsec:OnlineSurvey}

We conducted the online survey on Pollfish\footnote{\url{https://www.pollfish.com}} and recruited participants on a voluntary basis at an industrial fair, public spaces in Munich, and through social media. The survey begins with basic demographic questions regarding age and country of origin, then participants are presented with videos of 5-20 seconds (see Sec.~\ref{subsec:msg}), which they can replay if necessary. Each video presents, at random, one of the messages described in Section \ref{subsec:msg} modulated in one of the communication strategies $\mathcal{S}$. The order is randomized to counterbalance learning effects. Each video must be played through to the end, and is followed by two questions:

\textbf{\textit{I. ``What did you think the robot communicated?}} measures how the participants interpreted the presented message. Participants can choose between the following response options: \textit{\textbf{The robot [...]}} \textit{``tried to get attention''}, \textit{``operated normally''}, \textit{``communicated nothing''}, \textit{``announced to turn''}, \textit{``communicated that an error occurred''}, \textit{``will continue moving forward''}. \textbf{\textit{II. ``How difficult was it for you to understand the message the robot tried to communicate?''}} quantifies the difficulty of interpreting the message. The answer is given on a 5-point Likert Scale ranging from 1 (Extremely easy) to 5 (Extremely difficult), which has been Box-Cox-transformed and inverted for the data analysis (Section \ref{sec:results}). We include two checks to assess participants' attention levels, interspersed among the other questions. One asks participants to select "Strongly agree" on a 7-point Likert scale, and the other to provide a specific answer after watching a designated video. The final question of the survey asks about a preference for communication strategies: either ``Robot'' for $\mathcal{S}_{high}$ or ``LED'' for $\mathcal{S}_{low}$.

\subsection{Museum Experiment}\label{subsec:Museum}

\subsubsection{Location}
The in-person experiment takes place in the Science Communication Lab (SCL)'' of the ``Deutsches Museum'' in Munich, Germany, which offers an interactive space for temporary exhibits, educating the broader public about current trends in science and technology. Visitors are members of the public with a keen interest in technology.

\subsubsection{Experimental Design}

\begin{figure}[!t]
\vspace{0.2cm}
  \centering
  \includegraphics[width=0.9\columnwidth]{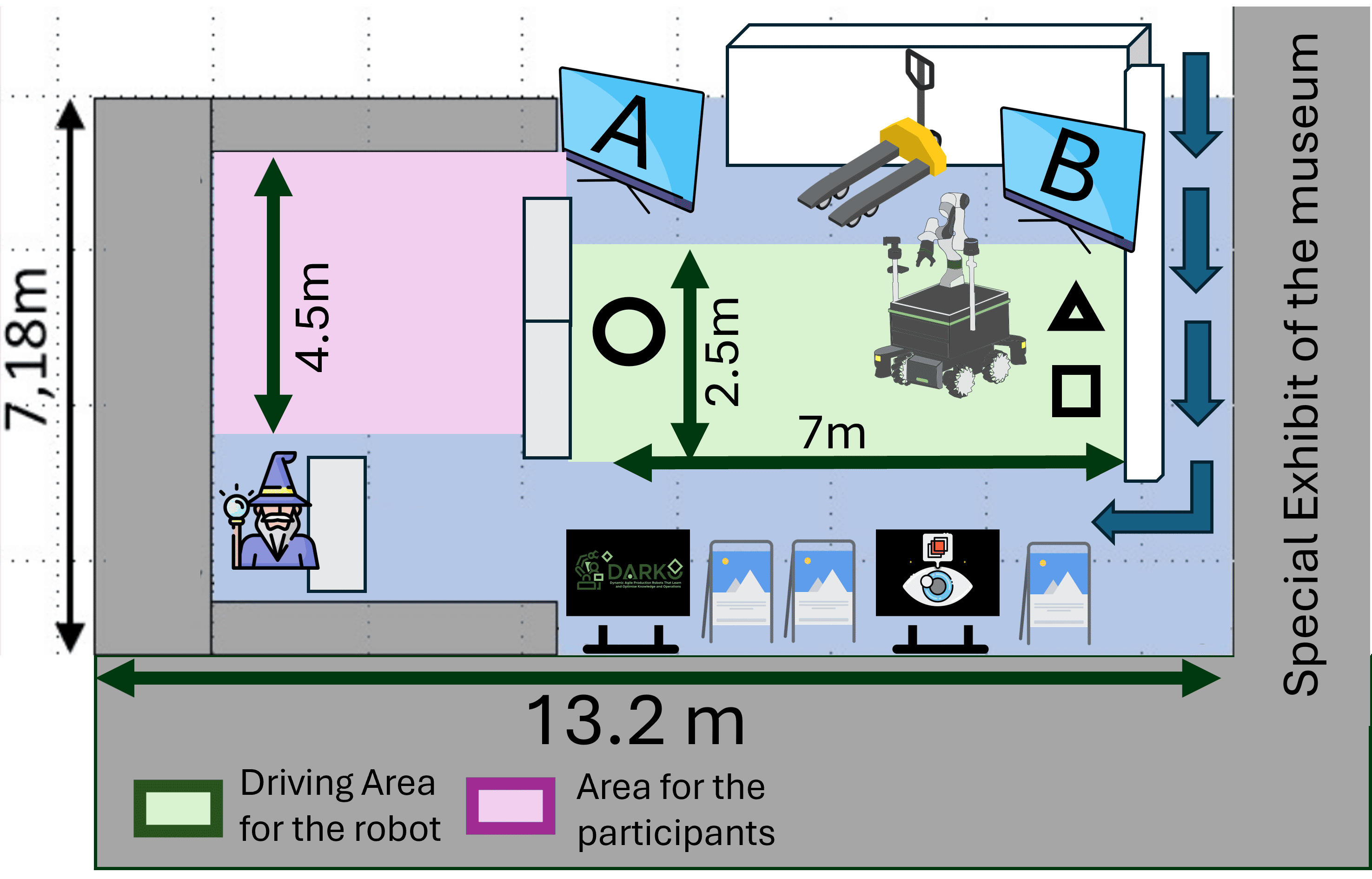}
  \caption{Setup of the experiment. Blue arrows indicate the flow of visitors. Geometric shapes show the starting positions of the robot: \textcircled{} for the error and attention indication, $\square$ for the turn indication, and $\triangle$ for the stuck behavior.}
  \label{fig:lab_setup}
\end{figure}

To address our research questions on the effects of communication strategies ($\mathcal{S}$) on the message legibility and intuitiveness, we replicate the online survey as closely as possible while adapting it to the conditions in the museum (see Figure~\ref{fig:frontal}). Our experimental setup is outlined in Figure~\ref{fig:lab_setup}. The experimental area (purple zone) is enclosed to minimize distractions from other museum visitors and ensure participant attention remains on the robot. Poster boards, room dividers, and TV screens are used for this purpose. The flow of visitors (indicated by blue arrows) is directed to minimize disturbances. In the online survey, the robot’s behavior was pre-recorded in a controlled laboratory environment. In contrast, public spaces present substantial challenges for robotic experiments due to their dynamic and unpredictable nature \cite{pelikan2024encountering}. To ensure robust operation and comparability across participant groups, we employ a Wizard-of-Oz (see the Wizard icon in Figure \ref{fig:lab_setup}) design that enables real-time control of the robot’s behavior. 

We collect participant responses in the museum using Mentimeter\footnote{\url{https://www.mentimeter.com/}}, replicating the online survey as interactive slides, including the attention checks. At the beginning of the experiment, participants are provided with a QR code to access the slides on their own devices. Alternatively, a paper-based version is also available. To ensure accessibility, the questions are displayed simultaneously on TV (A) (see Figure \ref{fig:lab_setup}). A second TV (B), reserved for the experimenter and wizard, shows only the current question number. 

\subsubsection{Experimental Procedure}

Our study spanned multiple days, with a session lasting approximately 30 minutes and serving groups of 5 to 30 participants. After participants receive the questionnaires (via Mentimeter or paper), we brief them on their right to withdraw and the purpose of the study (comparing robot communication strategies). After the demographic questions, they observed the messages displayed in random order (see Section~\ref{sec:methods}). Auditory cues marked the start and end of each message to maintain participant attention.

Between trials, the experimenter prepares the next condition by mounting or dismounting the NAO and administering questionnaires, while the wizard repositions the mobile robot according to instructions displayed on the TV (B). A handheld controller allows the wizard to steer the robot and trigger messages. The LEDs are manually toggled and remain active only during LED-based signaling conditions. Starting positions vary by scenario (see Fig. \ref{fig:lab_setup}). For the ``Error'' and ``Attention'' messages, the robot started at \textcircled{}. For the ``Turn'' message, the robot starts at $\square$, moves forward to stop between \textcircled{} and $\square$, then signals. In the crossing scenario of the ``Turn" message, the robot ($\square$) and pedestrian (\textcircled{}) move simultaneously upon an auditory cue, and the wizard stops the robot to signal the turn as their paths converge. Finally, for the ``Obstacle'' message, a manual forklift is placed at \textcircled{} between trials. The robot approaches the forklift from $\triangle$, and triggers the corresponding message.

\subsection{Participants}

We obtain written informed consent from all participants prior to both experiments, following consultation with local authorities. In our analysis, we exclude participants who failed both attention checks for both experiments. For the \textbf{Online Survey}, we recruit 100 participants, of whom we exclude three due to failed attention checks. Of the remaining $N = 97$ participants,  51.5\% are male, 43.3\% female, 2.1\% diverse, and 3.1\% preferred not to disclose their gender. The age ranges from 20 to 76 ($M = 36.2, SD = 13.0$). 86.6\% were from European countries (67.0\% from Germany), and 13.4\% were from other regions, such as Asia and North America. For the \textbf{Museum Experiment}, we recruit 139 participants (58.6\% male, 39.8\% female, 0\% diverse, and 1.6\% preferred not to disclose their gender). Ages range from 19 to 69 ($M = 39.8, SD = 12.5$). Of the participants, 82.2\% were from European countries (40.7\% from Germany), and 17.8\%  from other regions including Asia, North America, South America, Africa and Australia.



\section{Results} \label{sec:results}

\begin{figure*}[!t]
\vspace{0.2cm}
    \centering
    \includegraphics[trim= 0 28cm 0 0, clip, width=\textwidth]{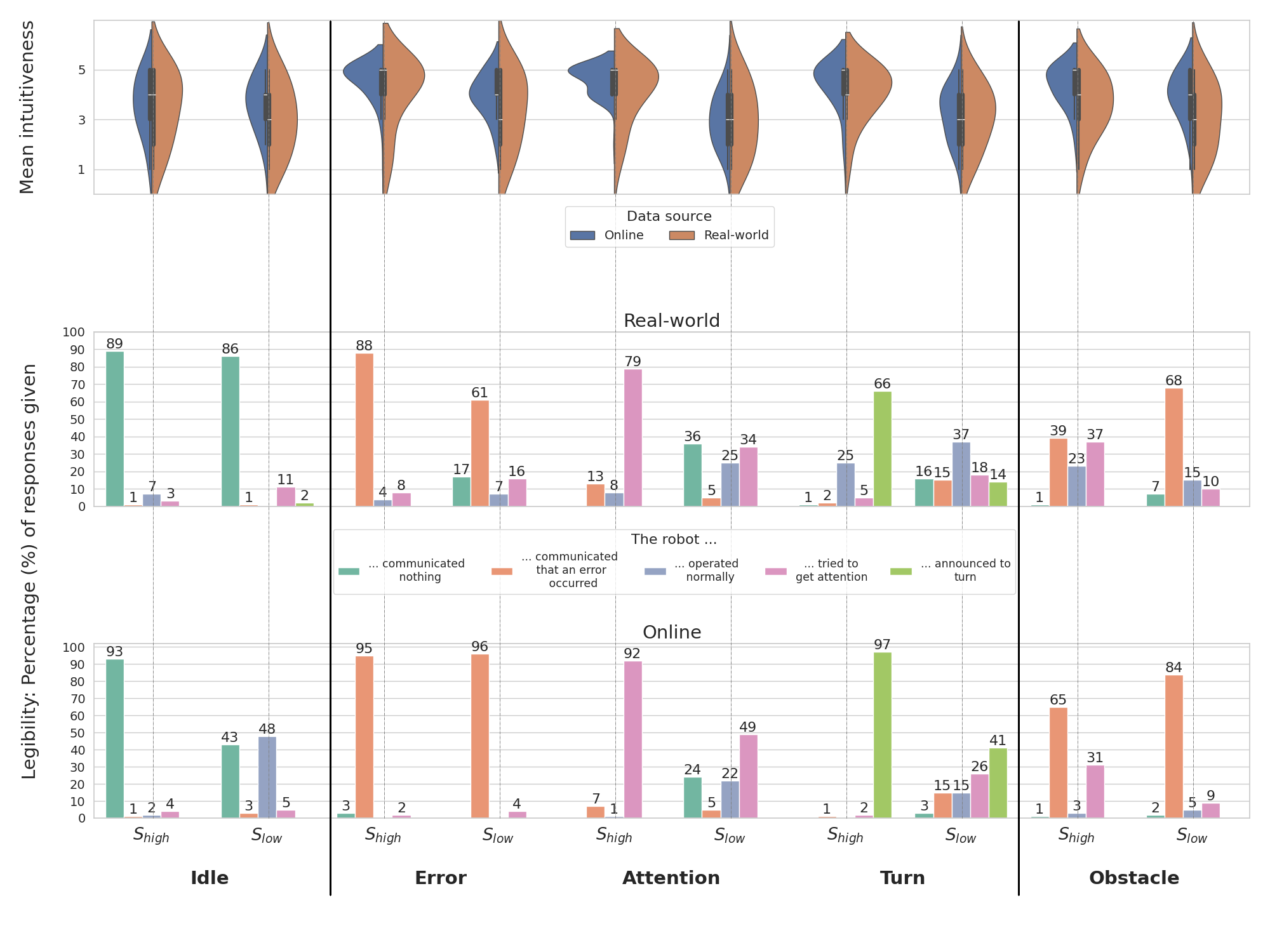}
    \includegraphics[trim= 0 2.7cm 0 12cm, clip, width=\textwidth]{figures/result_figure.png}
    \caption{Answers given by the participants in both experiments \textbf{Top:} Ratings regarding how difficult participants found it to interpret the messages (see Section \ref{subsec:msg}) (Intuitiveness), ranging from 1 - Extremely difficult to 5 - Extremely easy. \textbf{Bottom:} Legibility, i.e. to what extend could participants correctly resolve ``What did the robot communicate''?}
    \label{fig:result_plot}
\end{figure*}

In our studies, all messages were followed by the two questions outlined in Section \ref{subsec:OnlineSurvey} to measure perceived legibility and intuitiveness. We cluster responses into two separate metrics, which we will refer to as \textit{legibility score} and \textit{intuitiveness score} in the following. We calculate the legibility score as a ratio of the number of correct responses (set individually prior to the experiment) to all given responses. The intuitiveness score represents the central tendencies of the given replies to the second question on the Likert scale. Our metrics follow existing practices to evaluate participants' interpretations of robot communication \cite{lang2009feedback, fulton2022robot}. 

A higher score indicates higher legibility. The intuitiveness score we calculated is the Box-Cox-transformed ($\lambda = 2.02$) mean of the inverted scale across all questions with the same communication strategy, to account for the right-skew in the raw values. A higher score indicates a higher intuitiveness throughout the analysis. Both metrics are calculated separately for each communication strategy ($\mathcal{S}_{high}$ vs. $\mathcal{S}_{low}$) and for each participant. Overall, it is a 2x2 mixed design [(source of the data: online vs. real-world experiment; between-subjects; manipulated)  x (communication strategy: $\mathcal{S}_{high}$ vs.  $\mathcal{S}_{low}$; within-subjects)].

For each score, an ANOVA was calculated using the aforementioned independent variables. 
These analyses include the test for H1, H1.1 and H2. 
To test the other hypotheses, we performed $\chi^2$ tests of homogeneity on the legibility questions and Mann-Whitney U tests on the intuitiveness questions. In addition to the hypothesis tests, we examined the influence of speech on the legibility and intuitiveness of $\mathcal{S}_{high}$ by dividing it into two conditions: speaking and silence. The results for each question in detail are shown in Figure \ref{fig:result_plot}. 
All resulting \textit{p} values were corrected using the Bonferroni method to account for type 1 error accumulation.


\subsection{Hypotheses analyses}


The two-way ANOVA with the legibility score as dependent variable revealed significant main effects for the data source, $F(1, 404) = 262.83, p < .001$, partial $\eta^{2} = .39$ and communication strategy, $F(1, 404) = 53.97, p < .001$, partial $\eta^{2} = .12$. When it comes to the intuitiveness score, the two-way ANOVA indicated a significant main effect for the data source, $F(1, 404) = 21.16, p < .001$, partial $\eta^{2} = .05$, and for communication strategy, $F(1, 404) = 105, p < .001$, partial $\eta^{2} = .25$. Legibility was higher in the online survey ($M = 0.65, SD = 0.17$) than the real-world experiment ($M = 0.35, SD = 0.22$), and higher for $\mathcal{S}_{high}$ ($M = 0.55, SD = 0.20$), than for $\mathcal{S}_{low}$ ($M = 0.42, SD = 0.27$). The same applied to the intuitiveness: It was higher in the online survey ($M = 3.33, SD = 0.81$), than in the real-world experiment ($M = 2.95, SD = 1.03$), as well as higher for  $\mathcal{S}_{high}$ ($M = 3.55, SD = 0.84$), than $\mathcal{S}_{low}$ ($M = 2.70, SD = 0.87$). Those results support \textbf{H1.1} and \textbf{H2}. As for \textbf{H1}, Figure \ref{fig:result_plot} supports it by showing that the proportions of the responses differ between the online survey and the Real-world experiment.

\textbf{H2.1 (Supported):} A $\chi^2$ test for homogeneity shows a significant difference in the proportions between both communication strategies by legibility, $\chi^{2}(3, 409) = 122.91, p < .001, V = 0.55$. In 85\% of cases, participants interpreted the message as an attention request for $\mathcal{S}_{high}$ compared to 39\% for $\mathcal{S}_{low}$. Legibility was higher for $\mathcal{S}_{high}$ than for $\mathcal{S}_{low}$. For intuitiveness, $\mathcal{S}_{high}$ was significantly more intuitive ($M = 4.34, SD = 0.97$), than the $\mathcal{S}_{low}$ ($M = 2.97, SD = 1.22$), $U = 7856.0, p < .001, r = .31$. 

\textbf{H2.2 (Supported):} A $\chi^2$ test showed a significant difference between both communication strategies for the message communicating an error, $\chi^{2}(415) = 18.08, p < .001, V = 0.21$. In total, 91\% of participants interpreted the message as indicating an error when communicated using $\mathcal{S}_{high}$, compared to 77\% for $\mathcal{S}_{low}$, suggesting greater legibility for $\mathcal{S}_{high}$ than for $\mathcal{S}_{low}$. The Mann-Whitney U test revealed a significant difference in intuitiveness between the communication strategies, $U = 14050.0, p < .001, r = .29$, with $\mathcal{S}_{high}$ having higher intuitiveness ($M = 4.32, SD = 1.13$) than $\mathcal{S}_{low}$ ($M = 3.64, SD = 1.22$).

\textbf{H2.3 (Supported):} A $\chi^2$ test showed significantly different proportions of both communication strategies, $\chi^{2}(4, 781) = 179.34, p < .001, V = 0.48$. 77\% of participants perceived the message as a turn announcement for $\mathcal{S}_{high}$, whereas only 33\% did so for \textbf{$\mathcal{S}_{low}$}. As for intuitiveness, \textbf{$\mathcal{S}_{high}$} had higher ratings ($M = 3.71, SD = 1.72$), than$\mathcal{S}_{low}$ ($M=2.64, SD=1.50$), $U = 160217.0, p<.001, r=.46$. 

\subsection{Exploratory analyses}



We examined how speech within $\mathcal{S}_{high}$ affects the legibility and intuitiveness. A $\chi^2$ test revealed differences in the proportions of all three groups, $\chi^{2}(6, 511) = 102.94, p < .001, V = 0.32$. When including speech in $\mathcal{S}_{high}$, 54\% of participants classified the message as an error indication, compared to 28\% when speech was not included ($\mathcal{S}_{high}-Speech$). 77\%  of participants classified the message as an error for $\mathcal{S}_{low}$. The intuitiveness differed too, $H(599) = 57.65, p < .001, \eta^{2} = .09$, with$\mathcal{S}_{low}$ having the highest ($M = 3.98, SD = 1.08$), then $\mathcal{S}_{high}$ ($M = 3.49, SD = 1.25$) and then $\mathcal{S}_{high}-Speech$ ($M = 3.11, SD = 1.13$).



\section{Discussion}

\textbf{RQ1:}  Our results confirm \textbf{H1.1}, demonstrating that perceived communication effectiveness differs between online and real-world settings. Specifically, perceived legibility decreased significantly when transitioning from an online context to an in-person experiment. $\mathcal{S}_{low}$, our low-expressivity LED-based communication strategy, shows this effect most pronouncedly. Real-world environmental factors such as ambient noise, visual distractions, and varying viewing angles (see Figure \ref{fig:frontal}) likely contribute to this reduction. In contrast, $\mathcal{S}_{high}$, our high-expressivity communication strategy, exhibited a smaller decline in legibility and intuitiveness, indicating greater robustness to environmental interference. In the idle state, $\mathcal{S}_{low}$ maintained similar levels of legibility and intuitiveness across both settings. Here, the robot's stationary physical presence appeared to provide sufficient contextual grounding for interpreting the messages, suggesting that visual cues remain effective for stable interaction dynamics. Overall, our findings align with prior work indicating that the impact of embodiment on perception is context-dependent rather than uniform \cite{esterwood2025virtually}. Our results highlight the need for combined online and in-person evaluations to capture scalable perception trends and context-specific effects.

\textbf{RQ2:} We explored this research question through \textbf{H2}, which hypothesizes that messages communicated with $\mathcal{S}_{high}$ are perceived as more legible and intuitive than those communicated with $\mathcal{S}_{low}$. The results support \textbf{H2} across the evaluated messages. Regarding \textbf{H2.1} (Attention), communication with a higher expressive bandwidth ($\mathcal{S}_{high}$) achieved significantly higher legibility and intuitiveness scores than $\mathcal{S}_{low}$. The use of explicit semantic cues, such as deictic gestures and verbal utterances, enabled participants to interpret attention requests more consistently, with less variability. In contrast, lower ratings and greater variability indicate greater ambiguity for the $\mathcal{S}_{low}$ messages. Our findings suggest that communication strategies with greater expressive bandwidth provide clearer semantic grounding for critical interaction messages (e.g., path obstructions or attentional requests). In contrast, $\mathcal{S}_{low}$, which relies on abstract or learned mappings, appears to require additional contextual cues to achieve comparable interpretability  \cite{baraka2016enhancing, kalinowska2023embodied}. This implies a trade-off between semantic specificity and simplicity in the design of communication strategies.~\textbf{H2.2} (Error) was supported by the results. The $\mathcal{S}_{high}$ received statistically higher ratings for both legibility and intuitiveness than $\mathcal{S}_{low}$. A majority of participants correctly interpreted the error message under both communication strategies independently, indicating a high overall legibility for this state. For $\mathcal{S}_{low}$, successful interpretation can be attributed to strong cultural conventions associated with red visual signals, which are widely used to indicate warnings or error states across domains such as traffic signaling and consumer electronics \cite{baraka2016enhancing, akalin2025explain}. Exposure to these conventions occurs early in life, enabling predictable interpretation even in low-expressive communication contexts \cite{charisi2017children}. In contrast, $\mathcal{S}_{high}$ conveys error states through multimodal cues, which may increase clarity but can also introduce variability depending on how recipients weigh individual cues. This observation is consistent with a distinction between legibility and predictability \cite{dragan2013legibility}, where more expressive cues enhance legibility, while culturally grounded signals, such as red LEDs, support predictability through their grounding in established conventions.~\textbf{H2.3} (Turn): $\mathcal{S}_{high}$ achieved significantly higher legibility and intuitiveness ratings than $\mathcal{S}_{low}$. Specifically, 77\% of participants correctly identified the intention to turn when $\mathcal{S}_{high}$ was used, compared to 33\% for $\mathcal{S}_{low}$. Intuitiveness ratings followed the same pattern. The reduced ratings for LED-based communication suggest that (our) messages, inspired by vehicle turn indicators, may lose clarity outside the original traffic context, as they are decoupled from the standardized conventions that informed their design. In contrast, $\mathcal{S}_{low}$ conveyed turn intent through verbal utterances supplemented by spatially grounded semantic cues, such as pointing gestures. The supplemental cues support a more reliable interpretation when explicit spatial disambiguation, like turning direction, is required.

Our findings align with prior work, that navigation-related communication benefits from semantically explicit and spatially anchored cues, particularly in shared spaces \cite{halilovic2024exploring, schreiter2023advantages}. At the same time, they highlight the context-sensitivity of (light-based) signals with low-expressivity and the challenge of transferring vehicle-centric eHMI concepts to mobile robots, which requires proper adaptation \cite{fernandez2018passive, domonkos2020led, angelopoulos2022familiar}. Our results indicate that communication strategies with higher expressive bandwidth ($\mathcal{S}_{high}$) are more robust to convey directional intent in dynamic contexts.

Our exploratory analysis examined the contribution of speech as an individual channel in $\mathcal{S}_{high}$ compared to $\mathcal{S}_{low}$ in the obstacle scenario. While for $\mathcal{S}_{low}$ the message was correctly interpreted as an error state, in the $\mathcal{S}_{high}$, the removal of speech led to occasional ambiguity, with participants misinterpreting the remaining nonverbal cues as attention requests rather than error signals. This observation suggests that non-verbal expressive cues alone may require additional interpretive effort, particularly when multiple meanings are plausible \cite{dragan2013legibility}. In contrast, light-based signals seem to support the interpretation of system states, such as errors \cite{baraka2016enhancing, kalinowska2023embodied}.

\section{Conclusion and Future Work}

Through a two-stage evaluation in which participants experienced the messages of the robot online and in person, our study compares the legibility and intuitiveness of two communication strategies $\mathcal{S}$ with different expressive bandwidths for a mobile robot. We found that a unimodal communication strategy with low expressive bandwidth ($\mathcal{S}_{low}$) is equally effective for conveying simple, safety-critical states, such as errors, as a multimodal communication strategy with higher expressive bandwidth leveraging multiple social robotic cues ($\mathcal{S}_{high}$). On the other hand, the latter was shown to significantly enhance legibility and intuitiveness for complex, spatially dependent messages, such as turn announcements. We observe an overall decrease in both metrics during real-world trials compared to the online context, highlighting a virtual-to-real gap and the need to validate newly designed robot communication concepts with participants through both online and physical experiences. These results advocate a hybrid design paradigm that combines low-expressivity cues for system status with high-bandwidth interaction strategies. Since $\mathcal{S}_{low}$ and $\mathcal{S}_{high}$ differ along several dimensions (modalities, speech, gaze, pointing), we compared two holistic strategies at the bandwidth extremes rather than isolating bandwidth as a single factor. We leave a factorial study of the individual $\mathcal{S}_{high}$ modalities to future work.



\bibliographystyle{IEEEtran}
\bibliography{RO-MAN26/bibliothekl}

\end{document}